# Transcending the Attention Paradigm: Representation Learning from Geospatial Social Media Data

**Nick DiSanto, Anthony Corso, Benjamin Sanders, Gavin Harding**
**California Baptist University**
**Email: {nicolasc.disanto, acorso, bsanders, gavinjames.harding}@calbaptist.edu**

**Abstract**

While transformers have pioneered attention-driven architectures as a cornerstone of language modeling, their dependence on explicitly contextual information underscores limitations in their abilities to tacitly learn overarching textual themes. This study challenges the heuristic paradigm of performance benchmarking by investigating social media data as a source of distributed patterns. In stark contrast to networks that rely on capturing complex long-term dependencies, models of online data inherently lack structure and are forced to detect latent structures in the aggregate. To properly represent these abstract relationships, this research dissects empirical social media corpora into their elemental components, analyzing over two billion tweets across population-dense locations. We create Bag-of-Word embedding specific to each city and compare their respective representations. This finds that even amidst noisy data, geographic location has a considerable influence on online communication, and that hidden insights can be uncovered without the crutch of advanced algorithms. This evidence presents valuable geospatial implications in social science and challenges the notion that intricate models are prerequisites for pattern recognition in natural language. This aligns with the evolving landscape that questions the embrace of absolute interpretability over abstract understanding and bridges the divide between sophisticated frameworks and intangible relationships.

## 1   Introduction

The emergence of transformers has catalyzed a paradigm shift in the field of Natural Language Processing (NLP), ushering in an era of attention-driven frameworks. After sufficient pretraining and self-supervised learning on vast amounts of controlled data, these Large Language Models (LLMs) excel in task-specific environments, even as few-shot or zero-shot reasoners (Brown 2020; Kojima et al. 2023). However, the swift embrace of LLMs has been all-encompassing and overly optimistic, increasing the risk of impulsive applications and thoughtless trust in their outputs. The effectiveness of LLMs as conversational agents relies on the availability of structured data, monotonous human feedback, and meticulous prompt engineering. While LLMs demonstrate remarkable conversational aptitude, these fine-tuned training methods are largely domain-specific and starkly contrast the tacit nature of general human learning.

Natural intelligence emerges not from training on precise and composed corpora with a predetermined goal and a defined metric of success. Instead, general reasoning is an unsupervised synthesis of abstract connections between underlying patterns. However, industry-oriented architectures are notorious for being benchmark-driven (Sawada et al. 2023; Fu et al. 2023) to ensure apparent progress rather than adopting general-purpose learning practices that have intangible applications. This leads to some of the most advanced models struggling to make simple logical jumps that are trivial to human intuition (Efrat, Honovich, and Levy 2022) or generalize to environmental changes (Valmeekam et al. 2023). The inability of these models to discern intangible relationships is substantively attributable to the statically structured training data available. This necessitates the exploration of data sources that are implicitly imbued with subtle contextual nuances.

This study seeks to transcend heuristic benchmarking by investigating the underlying context of social media's latent patterns. Online data is specifically targeted because its variability restricts its embodiment of consistent relationships. This ensures that relationships are established in the aggregate over a general-purpose corpus, rather than being hand-picked as a valuable feature. Tweets, for instance, are characterized by their brevity and unpredictability, which present challenges in training deep models. As Mandal et al. (2022) point out, without transfer learning from pre-trained models, mapping to unencoded social media token sets can be extremely messy. However, the high volume and empirical association of user-generated online data make it an ideal candidate for fundamentally representing natural communication across diverse locations.

Additionally, language, by its very nature, serves as a robust empirical metric, as it primarily functions to communicate information about the world (Brown and Lenneberg 1954). Social media's volume and accessibility take this empirical application to the next level and make it a literal embodiment of human language at its most raw and unfiltered state. This investigation chose Twitter (now "X") content as its data source, aiming to gain tangible insights by deconstructing the data and examining geospatial correlations. More specifically, this analysis seeks to establish:

1. Whether Twitter communication styles can be represented as a function of geographic location
2. How similarity trends are indicative of regional relationships
3. The extent to which low-level word embeddings can embody high-level empirical context

These conclusions will determine the extent to which meaning can be derived from text itself, independent of the capabilities of a sophisticated model. Additionally, the results of this study seek to reveal the depth of geospatial language modeling and whether its variability increases with stochasticity. Additionally, if locational patterns can be mapped across Twitter even despite its arbitrary nature and inherent noise, it could warrant a paradigm shift toward indirect learning methods that prioritize the underlying context of real-world data over the overly explicit fine-tuning methods demanded by LLMs.

## 2  Related Works

Social media has become an increasingly common data source for language analysis, with broad applications ranging from academic sentiment analysis (Lasri, Riadsolh, and Elbelkacemi 2023) to evidence-based policy-making (Labafi et al. 2022). For example, Alotaibi et al. (2020) leverage social media data as a healthcare tool to identify trends of prevalent diseases in Saudi Arabia, offering insight into the interplay between online engagement and healthcare. Similarly, Rodrigues et al. (2021) pursue a broad temporal analysis of Twitter trends to gain insights into the efficiency of evaluating social media data in real-time. Social media data has also been used as a fine-tuning tool for LLMs, such as BERTweet (Nguyen, Vu, and Tuan Nguyen 2020), which achieves state-of-the-art performance on a variety of benchmarks. However, the specificity of this framework prompts a consideration of whether simpler models, when provided with sufficient data, can still find thematic patterns.

Extensive previous research has also used geographic location to explore the predictability of online data. For instance, Cheng et al. (2010) built a model capable of accurately estimating the location of microbloggers by relying on local word identification. Unique approaches have also inferred locations at varying granularities, starting with time zones and slowly narrowing down to specific zip codes (Mahmud, Nichols, and Drews 2014). This allows a hierarchal classification process to avoid early overfitting, focusing instead on high-level communication patterns. Locational analysis can also be multidisciplinary, with one study demonstrating that subjects of drug use and HIV outbreaks can often be triangulated to specific population-dense regions (Cai et al. 2020). Chandra et al. (2011) adopt a spatial reference framework that focuses exclusively on user interactions. They demonstrate that simple patterns and low-level features can still yield accurate models.

In order to establish the necessity—or lack thereof—of deep models as pragmatic predictors, the success tradeoff between powerful systems and contextual data must be evaluated. This contentious relationship is discussed by Halevy et al. (2009), who argue that the lack of concise solvability in NLP necessitates potentially noisy data that can be used to build refined networks. This preference for a robust corpus is also evident in LLM hallucinations, which occur when a model is unable to properly comprehend complex data (Lee 2023). Meanwhile, Ellis (2002) argues that human-level language modeling is entirely implicit and that the further models are abstracted, the more broadly beneficial they become. Strang et al. (2018) address the overabundance of complex implementations and their respective benchmarks by comparatively analyzing simple linear classifiers and intricate non-linear models. They aimed to identify the necessity of state-of-the-art systems and found that, while non-linear models are often advantageous, there are many applications in which they are overkill. This evidence suggests that simple models possess the capacity to uncover fundamental textual patterns, necessitating this analysis, which will gauge their practical applicability.

## 3  Methodology

**Dataset**

This study opted to use Twitter as its data source since it is publicly available in extremely high volume, utilized by diverse communities that build a comprehensive representation, and is frequently associated with measurable phenomena. From 2016 to 2020, an extensive dataset of approximately 2.5 billion publicly accessible tweets was gathered from various geographic regions using the standard Twitter API. These tweets were represented in JSON format, containing metadata of a variety of data types, such as author information and timestamps. This study opted to solely focus on the textual content and location, a decision that anonymized every individual tweet. However, additional features may be relevant to future studies, as discussed in the Future Work section. The general objective of the data collection was twofold:

- To accumulate a vast array of tweets from diverse locations across the United States
- To optimize the data's empirical applications

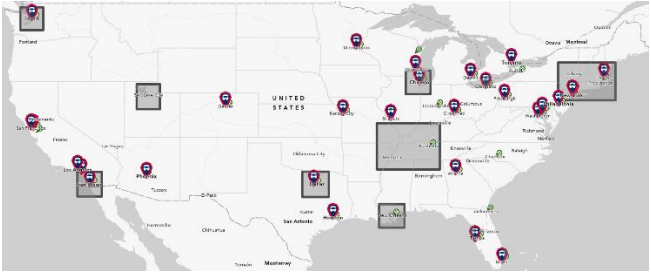

Figure 1: Data Collection Locations (Pins) and General Regions (Boxes)

Seventy specific data sites were identified, primarily centered around densely populated landmarks, such as sports stadiums and universities, and other generally populous regions with high activity. These locations were chosen to optimize the likelihood of empirical association with actionable events. Each site was defined by a latitudinal and longitudinal boundary, with the dimensions varying based on the rate of change in population density.

The dataset of tweets was gathered from diverse areas of the United States to encompass a broad cross-section of individuals and unique events. This consideration is essential to ensure that the data is fully representative of the platform's evolving content. Additionally, the data was collected comprehensively and autonomously, guaranteeing the procurement of discrete tweets devoid of predetermined patterns. The data collection processes were designed to be largely unsupervised, embracing high-volume data over a curated corpus. While this may result in the extensive inclusion of nonsensical posts and noise, embracing the tweets' unfiltered and instinctual nature optimizes the prospect of them being empirically inspired. Additionally, an unsupervised data collection process from diverse areas establishes a dynamic spectrum of user interactions, making it a valuable technique for unbiased data analysis. Further considerations in the data collection process, such as anonymization, social or political bias, misinformation, and toxicity are discussed in the Ethical Considerations section.

**Data Preparation and Tokenization**

This study opted to employ simple models, with the goal of determining whether these embeddings can associate empirical patterns akin to attention-dependent architectures. To break the dataset into its simplest terms and ensure low-level associations, Bag-of-Words (BoW) embeddings were constructed from each city's respective corpus. These were represented as arrays that mapped individual tokens to their respective frequencies. BoW modeling has historically served as a favored approach for categorizing experimental applications (Zhang, Jin, and Zhou 2010) due to its aggregated representation. Namely, Molero et al. (2023)

```
{                                          {
  "created_at": "Thu Aug 3                   "created_at": "Tues June 11
              08:24:49 +0000 2017",                       18:11:29 +0000 2019",
  "id": 746705011862845928,                  "id": 746705011862845928,
  "text": "u ever think you're late          "text": "Today's performance is just
           for work but really you're                   the latest disaster for the
           an hour early?",                             @RedSox.",
  "user": {                                  "user": {
    "name": "the_cali_notion",                 "name": "sox4lifealways",
    "coordinates": {32.714, -117.16},          "coordinates": {42.398, -71.054},
    "followers": 142                           "followers": 211
  },                                         },
  ...                                        ...
}                                          }
```

**Algorithm 1** – Dataset Preparation
**Input:** $city\_tweets$: List of lists of tweets, mapped to each city
**Output:** $city\_bows$: List of normalized BoW maps, mapped to each city
1: **procedure**
2:    $city\_bows \leftarrow \{\}$
3:    **for** $city$ in $city\_tweets$ **do**
4:       $bow \leftarrow \{\}$
5:       **for** $tweet$ in $city$ **do**
6:          $tweet \leftarrow preprocess(tweet)$
7:          **for** $token$ in $tweet$ **do**
8:             $bow[token]$++
9:          **end for**
10:       **end for**
11:       $bow \leftarrow bow[sort(bow.vals)[:5000]]$
12:       $bow.vals \leftarrow bow.vals / len(city)$    # normalization
13:       $city\_bows[city] \leftarrow bow$
14:    **end for**
15:    **return** $city\_bows$
16: **end**

**San Diego**

| Token | Value |
|---|---|
| "work" | 0.0199 |
| "day" | 0.0142 |
| "morning" | 0.0126 |
| "summer" | 0.0119 |
| "coachella" | 0.0081 |
| "view" | 0.0076 |
| ... | ... |

**Boston**

| Token | Value |
|---|---|
| "latest" | 0.0355 |
| "redsox" | 0.0217 |
| "today" | 0.0183 |
| "recommend" | 0.0177 |
| "see" | 0.0155 |
| "anyone" | 0.0149 |
| ... | ... |

Figure 2: Data Preparation Pipeline

illustrate its capability to effectively map to social media data, showcasing its enduring relevance even in the age of non-linear models and mitigate the issue of imbalanced data. After the BoWs were assembled, traditional data preparation techniques were applied to the individual tokens using the Natural Language Toolkit (Bird, Loper, and Klein 2009), including removing common stopwords and punctuation, standardizing case sensitivity, and stemming/lemmatizing individual tokens. The analysis was also constrained to English tweets to prevent the model from overfitting to language prediction, which would compromise the predictability of the content of the text itself. The goal of the data preparation process was to optimize the number of distinct tokens in the corpora while retaining all pertinent information. To improve computational efficiency and eliminate outlier tokens, the models were then trimmed to the top five thousand token occurrences of each. Finally, the corpora were proportionally scaled to the respective cities' populations to accommodate variations in population and subsequent social media usage.

**Similarity Evaluation**

Once the city embeddings were assembled and scaled, they were transformed into vectorized maps to be evaluated against one another. Cosine similarity was chosen as the criterion of the correlation between each city's BoW, since its simplicity and interpretability yield actionable results. Additionally, it has long been considered a standard metric for textual correlation in relational evaluation (Gunawan, Sembiring, and Budiman 2018) and social media classification (Fócil-Arias et al. 2017).

Given $C$ is the array of city embeddings, the sum of cosine similarities between each pair of individual cities $c_i, c_j \in C \mid i \neq j$ was calculated to represent their relationship. In order to compare each of these individual pairs $c_i$ and $c_j$, given $n$ is the token set shared between them, their similarity $S$ is formalized as:

$$S(c_i, c_j) = \frac{\sum_k^n c_i[k] \cdot c_j[k]}{\sqrt{\sum_k^n (c_i[k])^2} \cdot \sqrt{\sum_k^n (c_j[k])^2}}$$

This calculated each location's normalized word vectors' similarity with every other location, resulting in all seventy cities each having unique lists of results. In order to understand these results geospatially, the Haversine distance $d$ between each location was additionally calculated as follows, given the cities' geocentric latitudes $\varphi$ and longitudes $\lambda$:

$$d(c_i, c_j) = 2R \arcsin\left(\sqrt{\sin^2\left(\frac{\varphi_j - \varphi_i}{2}\right) + \cos\varphi_i \cdot \cos\varphi_j \cdot \sin^2\left(\frac{\lambda_j - \lambda_i}{2}\right)}\right)$$

This resulted in a two-dimensional array, mapping each city to its own unique embeddings, its similarity to each other location, and its distance from each other location.

## 4 Results

In order to scale each city pair's relationship by geospatial relevance, the similarities are represented as a function $f$ of distance. That is, each distance and each similarity between location $i$ and every other location is plotted linearly. This analysis can be represented as:

$$R = \left\{ f(c_i) \mid \forall c_i \in C, \ f(c_i) = \left\{ (d(c_i, c_j), S(c_i, c_j)) \mid \forall c_j \in C, \ c_j \neq c_i \right\} \right\}$$

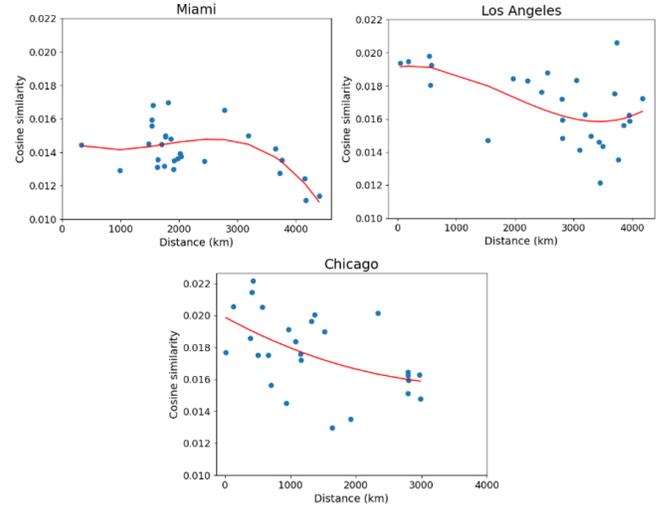

Figure 3: Examples of Similarity Functions for Individual Cities

The individual similarity functions displayed in Figure 3 are examples of $f(c_i) \in R$, where $R$ is the list of city distance maps. These specific maps were intentionally selected from far-apart locations to establish how the overall distribution can be understood by region. For example, $f(Los\ Angeles)$ and $f(Miami)$ demonstrate convincingly linear drops in similarity when compared with cities over 2,500 km away, barring outliers. This consistently disparate relationship with such cities implies that coastal cities are particularly unique in their representation. However, $f(Chicago)$ shows particularly high variance in its distribution, likely due to its midwestern location, which is equidistant to dozens of diverse cities of differing vernacular. Additionally, $f(Miami)$ shows a large cluster of unpredictable results around 2,000 km, likely due to the wide spectrum of far-away cities extending both North and West from Florida. This further emphasizes the divergence of vocabulary between cities of extreme distances or opposing coasts. The regional implications of these findings are further explored in the Discussion section.

While these individual correlative functions $f(c_i) \in R$ provide valuable feedback regarding the unique communication styles of different regions, it is also necessary to assess the functions as a whole. To do so, a composite function $|R| = \sum_i^n f(c_i)$ was established as an aggregation of the individual functions. This function is a valuable resource for understanding the general similarity correlation of the country, independent from specific regional variables. It also forms a more comprehensive view of the study's overall results.

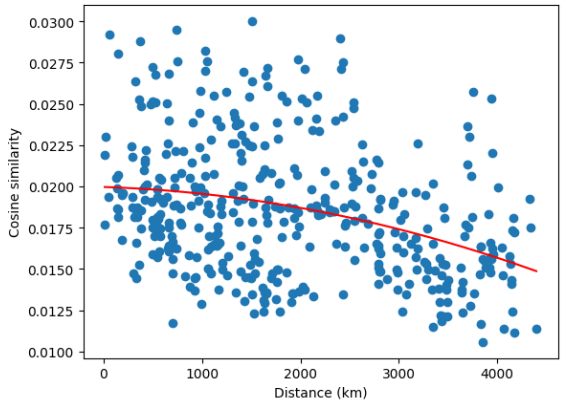

Figure 4: |R|: Aggregated Similarity Function

| Distance | Similarity | Δ S | ∑Δ S |
|---|---|---|---|
| 0 km | 0.0201 | 0% | 0% |
| 1,000 km | 0.0193 | 4% | 4% |
| 2,000 km | 0.0186 | 4% | 8% |
| 3,000 km | 0.0176 | 5% | 13% |
| 4,000 km | 0.0157 | 11% | 24% |
| 5,000 km | 0.0132 | 16% | 40% |

Table 1: Tabular View of Cosine Similarity Change over Distance

The high variance in |R| is to be expected since it is an unweighted composite of the individual cities, which have varying scales of similarity depending on location (as discussed previously). Additionally, the low scale of similarity is unsurprising considering a vast majority of Twitter content is entirely unique and is independent of locational implications. Nevertheless, even with little structure and unweighted results, a negative trend is discernible across all the locations. Table 1 shows a tabular representation of |R|, showing the increasing rate of similarity drop-off with distance. In the table, the average drop-off in similarity for every 1,000 km is represented as $\Delta S = \frac{S[d-1,000km] - S[d]}{S[d-1,000km]}$.

As shown in the table, the average cosine similarity between two cities extremely close in proximity to one another is 0.0201 but drops below 0.014 when the cities are separated by over 4,000 km, a ∑ΔS value of over 40%. Additionally, the rate of change of ΔS increases with each distance increment, hinting at a polynomial relationship. It is also interesting to note that the aggregated function shows largely a consistent scale of similarity, hovering between 0.01 and 0.03 for every comparison of city pairs. This shows that, while some regions have communication styles that particularly differ from other large regions, no individual city is independently unique from its surroundings. The amalgamation of regions results in a fluid transition between cities and subtle nuances that can only be detected in low-level examinations. Overall, both the similarity functions $f(c_i) \in R$ as well as the aggregated function |R| establish that:

1. In the aggregate, the correlation of online communication methods decreases with distance
2. Changes in similarity are particularly evident in cities over 2,500 km apart, with equidistant central cities absorbing attributes from both counterparts
3. Simple linear models, such as Bag-of-Word embeddings, are capable of identifying empirical trends

## 5 Discussion

This analysis is a convincing indication that differing underlying language patterns exist between cities of considerable distances. It presents several points of discussion regarding the locational analysis of communication styles. One important consideration is that these distinctions are purely in a unigram context. Therefore, the results are not attributable to differences in sentence structure or other large-scale linguistic attributes. Rather, they are a synthesis of tokenized characteristics that add up to a large-scale distribution. Additionally, BoW distinctions are a strong suggestion of empirical influence, because if every city generally communicates similarly, a sufficiently large BoW would normalize locational distinctions through raw computation.

Practical takeaways from this study include the implications of the increasing rate of ΔS after the distance passes 2,000 km in both $f(c_i) \in R$ and |R|. While similarity often clusters before this point, as $f(Miami)$ demonstrates, results become much more compelling as distance surpasses this threshold. In fact, many individual similarity functions show apparent linear rates of ΔS at these distances, suggesting that their representations are unique enough to differ equally from several other diverse locations. This is specifically seen in coastal cities, implying that locations that are not landlocked by other influential areas are entirely distinctive from cities of considerable distance.

Additionally, as previously mentioned, midwestern cities like Chicago exhibit the most variance in their results. This is likely because they are roughly equidistant to most other comparative locations and exhibit linguistic patterns that mix various geographic styles. These discoveries support the results of Kamath et al. (2012), which find that content similarity clusters between small distances but drops significantly after 3,000 km. Social science research has also affirmed these hypotheses, developing methods to segment specific regions of America and showing the unique distinctions of different geographic areas (Balsamo, Bajardi, and Panisson 2019). Overall, the results of this analysis emphasize a similar locational conclusion: natural language differs increasingly between coastal cities of long distances, while southern and midwestern cities pick up subtle similarities between both communication styles.

It is also necessary to acknowledge the inevitable presence of noise within social media data. While most models rely on datasets that are curated to exclusively contain beneficial information, user-generated content is inherently erratic. Consequently, most tweets analyzed in this study lacked any discernible geographic association. While this arbitrary data necessitated a broader analysis, it also makes the results hold particular significance since intangible structures that represent a small subset are unearthed across the entire corpus. Although refining the corpora by removing nonsensical data would have undeniably enhanced the models' performances, it would also have sacrificed the study's primary objective. The emergence of distributions from arbitrary data strengthens the argument for the existence of inherent context, even when employing simple metrics.

These findings also underscore a critical facet of employing large-scale NLP systems: the tradeoff between model interpretability and performance. The abstract relationships within these Twitter datasets remain largely enigmatic due to this study's emphasis on identifying pragmatic correlations, rather than understanding the black-box nature of such patterns. This focus aided data analysis since explainable machine learning often requires reducing data complexity and potentially higher-level associations, but it provides only presumptions of the real-world implications. Future studies that pinpoint specific features inherently associated with regional trends may yield more applicable results for the field of communication science and empower researchers to build a model that is just as interpretable as it is high-performing.

## 6   Future Work

This research explores the viability of traditional models to find complex locational patterns in noisy data. This has numerous implications for both the applicability of linear models and the development of advanced future archetypes. While BoW models and other low-level representational models can be contextually mapped to observable outcomes, additional work is necessary to establish the limitations of these findings. More specifically, it is vital to recognize the data preprocessing methods that were employed for optimization. While an important conclusion of this study is that unstructured, oftentimes meaningless data can show correlative results, truly tacit comprehension does not have this luxury; noise must be automatically filtered out at an extremely high level. Future implementations should further establish the performance cutoff due to increased noise, demonstrating when the model can no longer find pragmatic correlations.

An important consideration for future social media analysis is the consideration of hashtags, hyperlinks, and other nonalphanumeric-reliant text. This study opted for simplicity, removing all the tweets' symbolic characters from the start. However, some of these symbols are undoubtedly practical, yielding the opportunity to further establish locational distinctions. For example, Gupta et al. (2020) establish an automation process to derive semantically relevant hashtags and classify them based on empirical and domain-specific significance. Individual trendy hashtags have also been targeted to explore their respective political sentiments and demonstrate what members of the public web communicate in predictive ways (Teodorowski et al. 2022). Future work that accounts for specific symbol combinations could shed light on new methods for geospatial mapping and find additional relationships across online communities.

## 7   Conclusion

This study ventures beyond the conventional boundaries of language paradigms by exploring the nuanced relationship between geographic location and online communication. This has unveiled compelling evidence of distinct linguistic patterns emerging across diverse locations. The results of the study found that with distance, the similarity of communication methods consistently drops. Additionally, users from various general regions exhibit unique data representations, yielding fascinating empirical implications. This comprehensive examination provides a fresh perspective on how text can be considered not just expressions of thought, but also a reflection of context. Additionally, the simplicity of Bag-of-Word embeddings and unstructured data underscores the potential for uncovering hidden correlations within the chaotic realm of social media.

Finding geospatial correlation in data with no apparent structure demonstrates that models of minimal complexity can still learn implicitly and find subtle patterns. As a result, this research advocates for recognizing text as a profound source of representational patterns and abstract empirical features. As modern frameworks continue to grow in complexity and computational power, simple representations should not be underestimated. Embracing a temporary respite to a more primitive approach challenges a reconsideration of the necessary criteria for general-purpose intelligence. This also prompts a consideration of the viability of indirect learning methods that prioritize contextual understanding over explicit information. Implicit patterns found within text are a testament to the depth of language and the potential for future discovery within the ever-expanding world of data analysis.

One can only marvel at the possibilities if state-of-the-art models embrace an emphasis on intangible understanding over mere interpretability. Delving into the intricacies of how social media platforms can capture the nuances of human interaction promises to extend the frontiers of both communication science and NLP. Scaling architectures down to their core can make human intuition more computationally interpretable, yielding a distinction between the significance of non-linear models and the underlying context of rich empirical data. Striving for a harmonious blend between structured networks and amorphous relationships represents the ultimate objective in developing an agent capable of abstract reasoning. Such an implementation would have the potential to continue unraveling human data to fundamentally understand it at a high level. This process could unveil unique latent communication patterns that teach us more about human nature, online communication's role in representing cognition, and efficient future methods for further analysis.

## Ethical Statement

This study's results were exclusively an aggregation of billions of publicly accessible tweets. Since this data was collected entirely arbitrarily to enhance empirical association, there was no possibility of implicitly biased data selection. Additionally, since only the textual content and location were utilized as correlative features, the data was anonymized prior to the analysis. Keeping the corpus fully anonymous and emphasizing data volume instead of manual supervision eliminated the possibility of individual user identification. However, it is important to recognize that analyzing an autonomously gathered and unsupervised online corpus poses a potential threat of misinformation, social or political bias, and an insufficient representation of diverse communities. Additionally, social media's prominence and accessibility present an inherent risk of toxic or offensive content. While these issues did not appear relevant to the results of the study and were likely mitigated by sheer data volume, future research should compare the current locational correlations with those of a filtered dataset. This would establish the influence of social representation in aggregated locational correlations. Given these considerations and the study's emphasis on high-level regional correlations instead of individual tweet analysis, we do not foresee strong ethical concerns induced by our work.